\def\year{2021}\relax
\documentclass[letterpaper]{article} 
\usepackage{aaai21}  
\usepackage{times}  
\usepackage{helvet} 
\usepackage{courier}  
\usepackage[hyphens]{url}  
\usepackage{graphicx} 
\usepackage{natbib}  
\usepackage{caption} 
\usepackage{amssymb}
\usepackage{amsmath}
\usepackage{gensymb}

\title{Machine Learning aided Crop Yield Optimization}
\author{
    Chace Ashcraft,  Kiran Karra
    \\
}
\affiliations{
    Johns Hopkins University/Applied Physics Lab\\

    7701 Montpelier Rd\\
    Laurel, Maryland 20723\\
    kiran.karra@jhuapl.edu

}

\begin{document}

\maketitle

\begin{abstract}
We present a crop simulation environment with an OpenAI Gym interface, and apply modern deep reinforcement learning (DRL) algorithms to optimize yield.  We empirically show that DRL algorithms may be useful in discovering new policies and approaches to help optimize crop yield, while simultaneously minimizing constraining factors such as water and fertilizer usage. We propose that this hybrid plant modeling and data-driven approach for discovering new strategies to optimize crop yield may help address upcoming global food demands due to population expansion and climate change.
\end{abstract}

\section{Introduction}
Machine learning is enabling rapid progress in many areas of the physical and life sciences. However, its success depends on the availability of a large amount of data. For example, in computer vision, many databases of images \cite{mnist,cifar,imagenet,svhn} have spurned innovation both within that field, and machine learning in general. Similarly, large datasets in human language technologies (HLT), including IMDB \cite{imdb} for text, and LDC \cite{ldc} for speech, have had similar effects on both the subfield and machine learning in a broader context.  While these core subfields of machine learning have benefited from such large databases, the same cannot be said for machine learning aided agriculture.

Under many projections, Earth's population is projected to increase over the next few decades \cite{desa2019world}, and climate change is expected to adversely affect the way food is grown \cite{epa2017climate}. The application of machine learning to help accommodate for the projected population growth as well as adverse crop growing conditions through plant growth has been explored by both academia and industry.  Precision agriculture seeks to address this, in part, by providing real-time data driven analytics to help farmers decide when to apply fertilization, water, and other necessities, or use computer vision algorithms to determine critical time points, for example when to harvest fruits or vegetables.

A less explored avenue, and one which concerns this manuscript, is active control. More specifically, we are interested in the scenario where machine learning operates in a feedback loop to dynamically control and optimize an agricultural system in operation.  An example of this is the autonomous greenhouse challenge \cite{agc_1, agc_2}.  Here, the idea is to develop algorithms which monitor and control various greenhouse environmental aspects, including ventilation, irrigation, and temperature control, while maximizing crop health and yield.  However, the aforementioned challenge uses an expensive greenhouse setup (provided by the competition sponsor), is not readily available to most researchers, and requires a full growing season to produce a single, small dataset that cannot be reused in ``replay mode'' (rewinded and altered -- a common practice in machine learning to explore the state space). 

An alternative approach is to use software to simulate environments in which algorithms can exert active control. Several simulators at varying levels of fidelity exist. For crop yield, the most notable one is the Python Crop Simulation Environment (PCSE), which is based on the LINTUL3 \cite{lintul3} and WOFOST \cite{wofost} plant models.  From a modeling perspective, PCSE has high fidelity, but its primary drawback is that all model parameters need to be provided a-priori.  In particular, common farming tasks, such as irrigation and fertilization must be preplanned from seed sowing to harvest.  The drawbacks of this from an algorithmic perspective are primarily that it is more difficult, for an ML algorithm to learn because of: 1) the lack of immediate feedback, and 2) the long time-horizon over which the algorithm needs to provide predictions. More recently, \citet{overweg2021cropgym} published CropGym, a reinforcement learning environment for growing wheat that utilizes the LINTUL3 \cite{lintul3} crop model from PCSE. Their environment demonstrates the feasibility of using DRL to maximize harvestable biomass while minimizing fertilizer application. The work discussed here is similar, but uses a different crop model, problem formulation, and most importantly, a higher resolution time step to allow more fine grained algorithmic control over the environment. For these reasons, we believe it is still of interest to the community. 

Given the necessary simulation environments and datasets, we believe that machine learning can help address the problem of increased demands on the agricultural system due to both human population growth and climate change.  We believe that a growth, similar to the computer vision revolution which followed the inception of ImageNet \cite{imagenet} is possible for machine learning aided agriculture.  As a first step towards this front, we propose an easily accessible plant simulation environment \footnote{Source code can be found at: https://github.com/iscoe/croprl}, which can be used by machine learning practitioners and researchers without background expertise in agriculture, to develop policies which optimize the crop yield. In particular, this environment enables easy application of modern reinforcement learning techniques. Preliminary results using RlLib's \cite{rllib} implementation of Proximal Policy Optimization (PPO) with default parameters suggest that DRL may be a successful method of learning useful control policies for precision agriculture. We hope this is a new direction in which both the DRL community and the precision agriculture community will take interest.

\section{Plant Model}
Crop models can be important tools for assessing the impact of climate change on crop production and yield. In this work, we utilize the SIMPLE crop model~\cite{ZHAO201997} which includes 13 parameters to specify a crop type, with four of these for cultivar characteristics. Inputs required for the SIMPLE crop model include daily weather data, crop management, and soil water holding parameters.  The model has several limitations, including the lack of response to vernalization and photoperiod effect on phenology. The model accounts for irrigation but ignores nutrient dynamics, an important component of precision agriculture. The model also makes assumptions about well drained soil, which can overlook the effects of over-watering on a crop.  Regardless, \citet{ZHAO201997} suggest that the model can be fairly accurate while being easy to implement. We use the SIMPLE model as a starting point for implementing a viable reinforcement learning environment simulating crop yield.  

\section{Reinforcement Learning}
\begin{figure}
    \centering
    \includegraphics[width=0.40\textwidth]{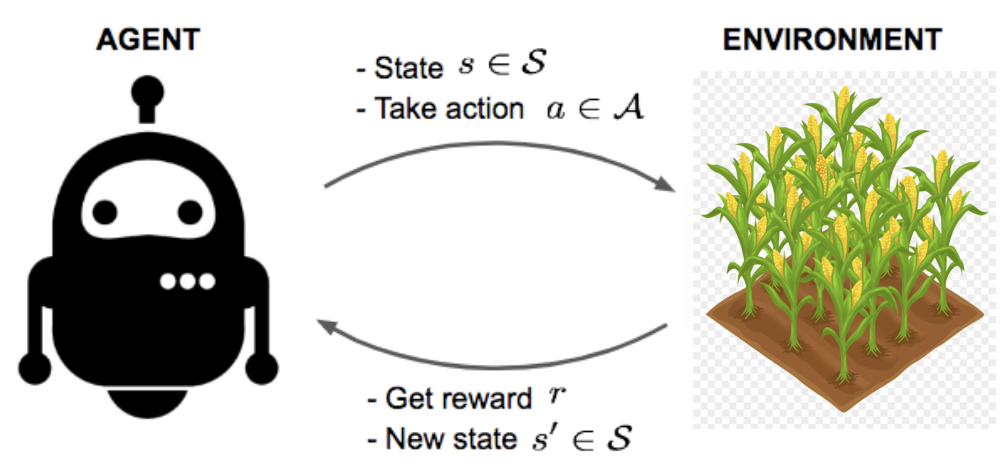}
    \caption{Reinforcement Learning's view of growing crops}
    \label{fig:rl_crops}
\end{figure}

Reinforcement learning is a subfield within machine learning, in which an agent seeks to interact with an external environment in order to optimize a given reward function. In reinforcement learning, the environment is formally specified as a Markov decision process (MDP) $\{S, A, T, R\}$, where $S$ is the state space, $A$ is the set of actions available to the agent, $T: S \rightarrow S$ is the transition function, and $R: S \times A \rightarrow \mathbb{R}$ is the reward function. Fig.~\ref{fig:rl_crops} depicts this in the context of growing crops, where the agent receives information regarding the state of the crops (for example, crop health, growth rate, etc.) and any rewards, and takes relevant actions (for example, irrigating, fertilizing, etc.) in the environment to acheive a user objective (for example, maximizing yield).

For this effort, we utilize an implemented version of PPO \cite{ppo} from the well established and vetted library \textbf{RlLib} \cite{rllib}.  PPO is a modern reinforcement learning algorithm which is known to work well ``out-of-the-box''.  PPO seeks to maximize the agent's reward by taking the largest possible improvement step on the policy during training, while minimizing catastrophic performance collapse often associated with large training steps.  This allows PPO to train quickly, while maintaining good performance.  In order to automatically take advantage of these reinforcement learning algorithms, we first wrap our SIMPLE plant environment into OpenAI's Gym interface \cite{gym}.  This is simply a standardized software interface for RL environments, and enables easy interfacing to RL algorithm implementations, including PPO.  


\section{Environment Description}


Our environment is configurable to use each of the parameters specified by the SIMPLE model. We set the time step of the environment to a single day, to match the SIMPLE model. At each time step, the agent receives an observation consisting of a noisy reading of plant's biomass, the cumulative daily mean temperature since sowing, the number of days since sowing, the day's weather (as given in Table~\ref{tab:weather_variables}), a noisy reading of the next day's weather, and a noisy reading of plant available water (PAW). We define noisy to mean sampled from a Gaussian distribution with the actual state value as the mean and a given standard variation. The agent action is a single continuous value in the interval $[0, 100]$ representing the amount of water to apply, in millimeters (mm), to the crop. The reward function is defined as follows:

$$
R(s, a) =
\begin{cases}
  C_I \times a + Y & \text{at harvest date} \\
  C_I \times a & \text{otherwise}
\end{cases}
$$

where $s$ is the current state, $a$ is the current action and $Y$ is the final yield in metric tons per hectare. $C_I$ is a hyperparameter representing the cost to irrigate, which we set to $C_I = -1.25 \times 10^{-4}$. The goal of PPO in this case is to maximize the reward, which is positively correlated with the yield, $Y$, and negatively correlated with the amount of water used, while accounting for the noisy state observations. 

\begin{table}[]
    \centering
    \begin{tabular}{c|c}
        TMAX & Daily maximum temperature (\degree C) \\ \hline
        TMIN & Daily minimum temperature (\degree C) \\ \hline
        TAVG* & Daily mean temperature (\degree C) \\ \hline
        RAIN & Daily rainfall (mm) \\ \hline
        SRAD & Daily solar radiation (MJ/m\textsuperscript{2}) \\ \hline
        CO2 & Atmospheric CO2 concentration (PPM) \\ \hline
        VAP** & Average daily vapor pressure (hPa) \\ \hline
        WIND** & Average wind speed at 2 meters (m/s) \\ \hline
    \end{tabular}
    \caption{Weather parameters used by our environment. \\ *Not specified as weather variable in SIMPLE, but used in our model. \\ **Not used in original SIMPLE model, but used to compute reference evapotranspiration in ours using the Penman-Monteith method~\cite{allen1998crop}.}
    \label{tab:weather_variables}
\end{table}

\section{Preliminary Results}


We use the parameters for Russet Potatoes from Benton, Washington, USA given in \cite{ZHAO201997}, weather data from Benton starting April 1st, 2000 \footnote{from \url{https://weather.wsu.edu}}, and standard deviations for weather parameters given in Table~\ref{tab:parameter_stds}. 

\begin{table}[]
    \centering
    \begin{tabular}{c|c}
                        \hline
        Biomass & 1 \\ \hline
        PAW & 1 \\ \hline
        TMAX & 3 \\ \hline
        TMIN & 2 \\ \hline
        TAVG & 2 \\ \hline
    \end{tabular}
    \quad
    \begin{tabular}{c|c}
                    \hline
        RAIN & 3 \\ \hline
        SRAD & 1 \\ \hline
        CO2 & 1 \\ \hline
        VAP & 1 \\ \hline
        WIND & 6 \\ \hline
    \end{tabular}
    \caption{Standard deviations used to add noise to observation variables to create the next day's weather forecast.}
    \label{tab:parameter_stds}
\end{table}

\begin{figure}
    \centering
    \includegraphics[width=0.4\textwidth]{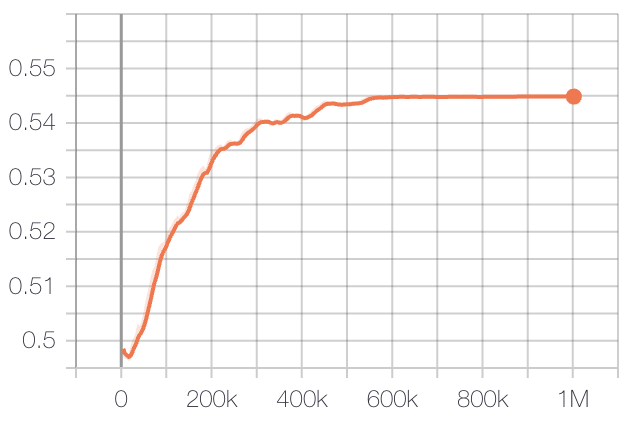}
    \caption{Learning curve for PPO agent growing Russet potatoes simulated with weather from Benton, WA. The $x$ axis shows training steps, and $y$ axis shows normalized average reward.}
    \label{fig:rl_mean_reward}
\end{figure}

Using RLlib's implementation of PPO with default parameters and default neural network architecture, we observe the mean reward growth curves shown in Figure~\ref{fig:rl_mean_reward}. The initial indications of learning are promising, but note that performance seems to plateau sub-optimally. Intuitively, optimal performance should be attained by spending the least on irrigation as possible while keeping the soil saturated. For example, using a constant action of $10 \frac{mm}{m^2}$ irrigation leads to net normalized reward of $0.976$ in the current implementation, while the PPO agents seems to plateau closer to $0.55$. 

\section{Future Work}
Given the very early nature of this project, there are many tasks future work could, or should, address. We propose two general directions we are interested in: crop modeling and advanced DRL.

For crop modeling, we should further validate the correctness and accuracy of our model using data from the field or real-world data collected from crop growing experiments. Validation is important for obtaining useful results for applied agriculture and understanding the DRL performance. We would also find it useful and worthwhile to further augment the model to account for other agent actions, such as the application of fertilizer and deciding when to harvest. Another useful improvement would be to improve the model to account for other realistic conditions, such as over-watering, or implement a more complex model entirely, such as those used by PCSE \cite{pcse}. Finally, finding and using more accurate simulation parameters, such as for daily weather, would also make our simulations more realistic. 

Future work in DRL could include the application of different algorithms to the simulator, attempting hyperparameter tuning, and experimenting with various neural network architectures. Each of these can significantly affect the performance of DRL on a task. Additionally, many other observation representations and reward functions are possible, some of which may facilitate faster learning. Lastly, because DRL tends to require significant amounts of data, and real-world data is typically difficult to obtain, we believe application of techniques such as offline reinforcement learning or meta-learning would be interesting to explore.


\section{Conclusion}
We put forth the idea that with relevant datasets and software simulation environments, the problem of supporting future growing populations under climate change can be possible and furthermore aided by machine learning. As a proof of concept for this idea, we showcase a simple plant simulation model with an OpenAI Gym interface, and train an agent using a modern RL algorithm to learn a policy for optimizing crop yield. We seek to continue this work by building better plant models to increase the fidelity of our simulation and make its policies more relevant to real-world conditions.  

\bibliography{main}

\begin{thebibliography}{19}
\providecommand{\natexlab}[1]{#1}
\providecommand{\url}[1]{\texttt{#1}}
\providecommand{\urlprefix}{URL }
\expandafter\ifx\csname urlstyle\endcsname\relax
  \providecommand{\doi}[1]{doi:\discretionary{}{}{}#1}\else
  \providecommand{\doi}{doi:\discretionary{}{}{}\begingroup
  \urlstyle{rm}\Url}\fi

\bibitem[{Allen et~al.(1998)Allen, Pereira, Raes, Smith et~al.}]{allen1998crop}
Allen, R.~G.; Pereira, L.~S.; Raes, D.; Smith, M.; et~al. 1998.
\newblock Crop evapotranspiration-Guidelines for computing crop water
  requirements-FAO Irrigation and drainage paper 56.
\newblock \emph{Fao, Rome} 300(9): D05109.

\bibitem[{Brockman et~al.(2016)Brockman, Cheung, Pettersson, Schneider,
  Schulman, Tang, and Zaremba}]{gym}
Brockman, G.; Cheung, V.; Pettersson, L.; Schneider, J.; Schulman, J.; Tang,
  J.; and Zaremba, W. 2016.
\newblock Openai gym.
\newblock \emph{arXiv preprint arXiv:1606.01540} .

\bibitem[{de~Wit(2021)}]{pcse}
de~Wit, A. 2021.
\newblock Python Crop Simulation Environment.
\newblock \urlprefix\url{https://pcse.readthedocs.io/en/stable/index.html}.

\bibitem[{Deng et~al.(2009)Deng, Dong, Socher, Li, Li, and Fei-Fei}]{imagenet}
Deng, J.; Dong, W.; Socher, R.; Li, L.-J.; Li, K.; and Fei-Fei, L. 2009.
\newblock Imagenet: A large-scale hierarchical image database.
\newblock In \emph{2009 IEEE conference on computer vision and pattern
  recognition}, 248--255. Ieee.

\bibitem[{DESA(2019)}]{desa2019world}
DESA, U. 2019.
\newblock World Population Prospects 2019. United Nations. Department of
  Economic and Social Affairs.
\newblock \emph{World Population Prospects 2019} .

\bibitem[{EPA(2017)}]{epa2017climate}
EPA, U. 2017.
\newblock Climate impacts on agriculture and food supply.
\newblock \emph{United States Environmental Protection Agency, Washington, DC}
  .

\bibitem[{Hemming et~al.(2019)Hemming, de~Zwart, Elings, Righini, and
  Petropoulou}]{agc_1}
Hemming, S.; de~Zwart, F.; Elings, A.; Righini, I.; and Petropoulou, A. 2019.
\newblock Remote Control of Greenhouse Vegetable Production with Artificial
  Intelligence—Greenhouse Climate, Irrigation, and Crop Production.
\newblock \emph{Sensors} 19(8).
\newblock ISSN 1424-8220.
\newblock \doi{10.3390/s19081807}.
\newblock \urlprefix\url{https://www.mdpi.com/1424-8220/19/8/1807}.

\bibitem[{Hemming et~al.(2020)Hemming, Zwart, Elings, Petropoulou, and
  Righini}]{agc_2}
Hemming, S.; Zwart, F.~d.; Elings, A.; Petropoulou, A.; and Righini, I. 2020.
\newblock Cherry Tomato Production in Intelligent Greenhouses—Sensors and AI
  for Control of Climate, Irrigation, Crop Yield, and Quality.
\newblock \emph{Sensors} 20(22).
\newblock ISSN 1424-8220.
\newblock \doi{10.3390/s20226430}.
\newblock \urlprefix\url{https://www.mdpi.com/1424-8220/20/22/6430}.

\bibitem[{Krizhevsky(2009)}]{cifar}
Krizhevsky, A. 2009.
\newblock Learning multiple layers of features from tiny images.
\newblock Technical report.

\bibitem[{LeCun and Cortes(2010)}]{mnist}
LeCun, Y.; and Cortes, C. 2010.
\newblock {MNIST} handwritten digit database
  \urlprefix\url{http://yann.lecun.com/exdb/mnist/}.

\bibitem[{Liang et~al.(2017)Liang, Liaw, Nishihara, Moritz, Fox, Gonzalez,
  Goldberg, and Stoica}]{rllib}
Liang, E.; Liaw, R.; Nishihara, R.; Moritz, P.; Fox, R.; Gonzalez, J.;
  Goldberg, K.; and Stoica, I. 2017.
\newblock Ray rllib: A composable and scalable reinforcement learning library.
\newblock \emph{arXiv preprint arXiv:1712.09381} 85.

\bibitem[{Maas et~al.(2011)Maas, Daly, Pham, Huang, Ng, and Potts}]{imdb}
Maas, A.~L.; Daly, R.~E.; Pham, P.~T.; Huang, D.; Ng, A.~Y.; and Potts, C.
  2011.
\newblock Learning Word Vectors for Sentiment Analysis.
\newblock In \emph{Proceedings of the 49th Annual Meeting of the Association
  for Computational Linguistics: Human Language Technologies}, 142--150.
  Portland, Oregon, USA: Association for Computational Linguistics.
\newblock \urlprefix\url{http://www.aclweb.org/anthology/P11-1015}.

\bibitem[{Netzer et~al.(2011)Netzer, Wang, Coates, Bissacco, Wu, and Ng}]{svhn}
Netzer, Y.; Wang, T.; Coates, A.; Bissacco, A.; Wu, B.; and Ng, A.~Y. 2011.
\newblock Reading Digits in Natural Images with Unsupervised Feature Learning .

\bibitem[{of~Pennsylvania(2021)}]{ldc}
of~Pennsylvania, U. 2021.
\newblock Linguistic Data Consortium.
\newblock \urlprefix\url{http://ldc.upenn.edu}.

\bibitem[{Overweg, Berghuijs, and Athanasiadis(2021)}]{overweg2021cropgym}
Overweg, H.; Berghuijs, H. N.~C.; and Athanasiadis, I.~N. 2021.
\newblock CropGym: a Reinforcement Learning Environment for Crop Management.

\bibitem[{Schulman et~al.(2017)Schulman, Wolski, Dhariwal, Radford, and
  Klimov}]{ppo}
Schulman, J.; Wolski, F.; Dhariwal, P.; Radford, A.; and Klimov, O. 2017.
\newblock Proximal policy optimization algorithms.
\newblock \emph{arXiv preprint arXiv:1707.06347} .

\bibitem[{Shibu et~al.(2010)Shibu, Leffelaar, Van~Keulen, and
  Aggarwal}]{lintul3}
Shibu, M.; Leffelaar, P.; Van~Keulen, H.; and Aggarwal, P. 2010.
\newblock LINTUL3, a simulation model for nitrogen-limited situations:
  Application to rice.
\newblock \emph{European Journal of Agronomy} 32(4): 255--271.

\bibitem[{Van~Diepen et~al.(1989)Van~Diepen, Wolf, Van~Keulen, and
  Rappoldt}]{wofost}
Van~Diepen, C.~v.; Wolf, J.; Van~Keulen, H.; and Rappoldt, C. 1989.
\newblock WOFOST: a simulation model of crop production.
\newblock \emph{Soil use and management} 5(1): 16--24.

\bibitem[{Zhao et~al.(2019)Zhao, Liu, Xiao, Hoogenboom, Boote, Kassie, Pavan,
  Shelia, Kim, Hernandez-Ochoa, Wallach, Porter, Stockle, Zhu, and
  Asseng}]{ZHAO201997}
Zhao, C.; Liu, B.; Xiao, L.; Hoogenboom, G.; Boote, K.~J.; Kassie, B.~T.;
  Pavan, W.; Shelia, V.; Kim, K.~S.; Hernandez-Ochoa, I.~M.; Wallach, D.;
  Porter, C.~H.; Stockle, C.~O.; Zhu, Y.; and Asseng, S. 2019.
\newblock A SIMPLE crop model.
\newblock \emph{European Journal of Agronomy} 104: 97--106.
\newblock ISSN 1161-0301.
\newblock \doi{https://doi.org/10.1016/j.eja.2019.01.009}.
\newblock
  \urlprefix\url{https://www.sciencedirect.com/science/article/pii/S1161030118304234}.

\end{thebibliography}

\end{document}